# DETECTING DRIVER FATIGUE WITH EYE BLINK BEHAVIOR

Ali Akin[1] and Habil Kalkan[2]

[1]Computer Engineering Department,
Gebze Technical University, Kocaeli, TURKIYE
aakin2021@gtu.edu.tr

[2]Computer Engineering Department,
Gebze Technical University, Kocaeli, TURKIYE
hkalkan@gtu.edu.tr

## ABSTRACT

*Traffic accidents, causing millions of deaths and billions of dollars in economic losses each year globally, have become a significant issue. One of the main causes of these accidents is drivers being sleepy or fatigued. Recently, various studies have focused on detecting drivers' sleep/wake states using camera-based solutions that do not require physical contact with the driver, thereby enhancing ease of use. In this study, besides the eye blink frequency, a driver adaptive eye blink behavior feature set have been evaluated to detect the fatigue status. It is observed from the results that behavior of eye blink carries useful information on fatigue detection. The developed image-based system provides a solution that can work adaptively to the physical characteristics of the drivers and their positions in the vehicle.*

## KEYWORDS

*Driver, Fatigue, Blinking, Head Movements, Face Landmark,*

## 1. INTRODUCTION

Traffic accidents occurring annually on roads worldwide cause approximately 1.2 million deaths and injure or disable 50 million people. These accidents also result in approximately 518 billion dollars in material damage each year [1]. According to the American Automobile Association, 16.5% of traffic accidents resulting in fatalities and 12.5% of those causing injuries are due to drowsy driving [2, 3]. In 2020, a total of 983,808 traffic accidents occurred in Turkey, with 833,533 resulting in property damage and 150,275 in fatal injuries. According to reports from the Turkish Statistical Institute, 88.3% of accidents are caused by driver errors [4]. Many studies have been conducted in the literature to detect driver fatigue, and various systems have been developed.

The human body transitions between NREM (Non-Rapid Eye Movement) and REM (Rapid Eye Movement) sleep phases (Figure 1). Each stages of NREM (N1, N2 and N3) sleep lasts between 5-15 minutes. At the N1 state, the heart rate and blood pressure decrease and the head and eye movement slows down [6]. In driver detection systems, it is crucial to detect the driver before the initial phase of sleep. Systems that continuously evaluate driver fatigue based on data obtained from the driver or the driving environment during the driving process are known as Driver Fatigue Detection Systems (DFDS).

The target of DFDS is to detect the user's sleep state during the NREM phase and provide warnings to take a break or stop driving to prevent potential accidents.

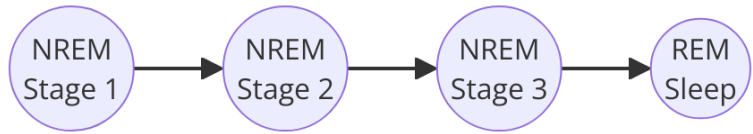

Figure 1. Sleep Stages.

On April 23, 2021, the European Union legislated the use of driver alert systems, making it mandatory for all new vehicles manufactured from 2022 onwards to include such systems [7]. Currently, these systems are known to be present in new vehicles, but there is a need for modular DFDS systems for vehicles produced before this date or those not covered by this legal regulation.

Due to the needs and the regulations, many studies have been performed to detect driver's fatigue using different features (visual and physical) obtained from the drivers during the drive. Visual features include eye and head movements, while physical features include ECG, internal and external temperature, lane violations, accelerator pedal usage, steering grip, and more. During the fatigue or transition to sleep stage, the heart rate decreases and body temperature changes. E. Rogado and his team [8] conducted fatigue assessments using ECG data obtained from the steering wheel and digital temperature sensors to measure internal and external temperature differences. However, this method may not be practical for drivers wearing gloves. Yogesh and his team [9] developed an active alert system by detecting yawning using image processing. Unlike Volkswagen's driver warning system, which activates at speeds of 65 km/h or more, this system is always active, as fatigue can cause accidents even in slow-moving traffic [8]. Hamzah et. al [10] detected fatigue by processing neural waves from the brain. However, this EEG-based system can be obtained with wearable technologies. Using helmet-like equipment continuously inside the vehicle can make the driver uncomfortable, leading to a disadvantage. In conclusion, the negative aspects of using physical features include the difficulty in obtaining data, the need for correct data readings from sensors on seat belts, and the necessity for equipment in EEG-based systems.

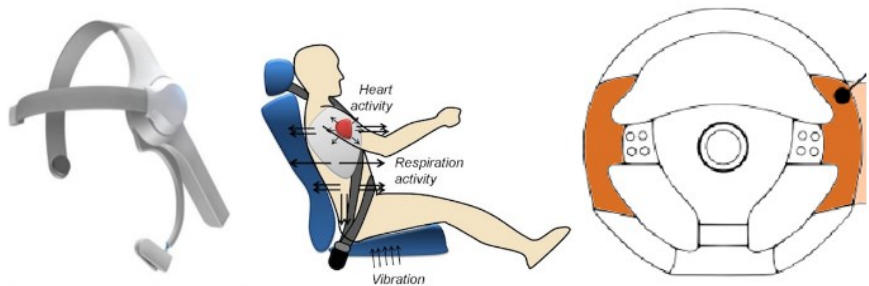

Figure 2. In-car sensor points for physical features [10] [11]

Visual features in driver fatigue detection systems include eye, mouth, and head movements. Systems primarily using visual features are mechanisms that work with an image-capturing device, evaluate the obtained images, and make fatigue decisions based on them. Burcu and Yaşar [12] proposed a real-time system for detecting driver fatigue. They used ConNN to detect fatigue based on features obtained from image processing. The features used were PERCLOS (Percentage of Closure Eye) and FOM (Frequency of Mouth). Data from 150 frames were collected for five-second segments, and PERCLOS and FOM were calculated. The YawDD dataset, consisting of 322 videos from different ethnicities and genders, showed a success rate of 98% [11]. However, the high success rate is attributed to the exclusion of data from drivers wearing sunglasses during testing and training.

Similarly, Revna et. al [13] developed a real-time driver fatigue detection system on Raspberry Pi 3 using image processing. In the images used to calculate PERCLOS, each frame was marked as open or closed eye. Ayşenur et. al [14] used the Raspberry Pi 4 and the Dlib library for driver fatigue detection. They used the Dlib library, which marks the face region at 68 points, to calculate eye openness.

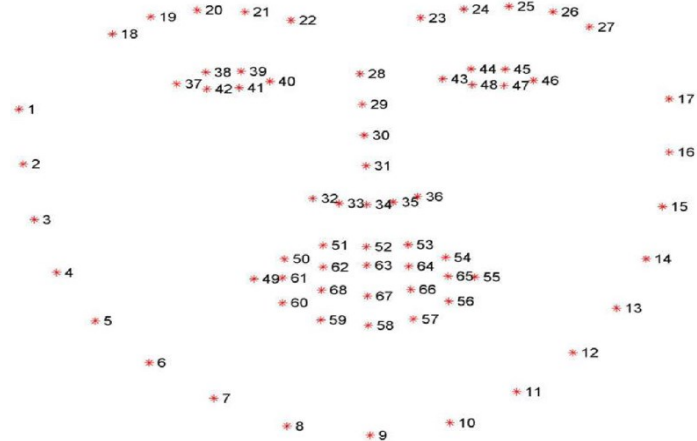

Figure 3. Dlib facial landmarks [15].

The PERCLOS in [13] was calculated by dividing the number of closed-eye frames within one minute by the total number of frames and multiplying by 60. If the PERCLOS value is greater than 80, the system detects drowsiness and alerts the driver. Mehmet et.al [16] used PERCLOS for driver fatigue detection. They classified eye open-closed status using a support vector machine and OpenCV. For the study, 2111 eye images were collected from 19 people and classified using SVM. Ali and Ergun [17] stated that PERCLOS is an effective feature in detecting driver fatigue. They detected eye open or closed status using infrared technology. Mariella et.al [18] extracted 35 features related to blinking behavior and head movements. They selected features using the K-NN method and found that the duration of reopen the eyes, the amplitude speed ratio of closing the eyes, the eyelid gap, and head dropping forward or backward are significant on fatigue detection.

Brand studies generally work based on the driver's movements. They analyze the driver's behavior and warn about signs of fatigue by considering steering movements, actions on other devices (signal lights, windshield wipers, etc.), and continuous driving time. In Volkswagen's fatigue detection system [19], driver behavior and irregular steering movements are continuously evaluated with sensors on the steering wheel at speeds of 65 km/h or more. In Ford's driving support systems [20], the lane tracking system is integrated to understand the driver's fatigue or distraction. A system has been developed that continuously monitors the driver's position and alerts the driver in case of swerving or deviation, informing them that they need to rest. Additionally, Ford's driving support systems provide more effective warnings with vibrations on the steering wheel.

The success of fatigue detection algorithms basically depends on evaluating the fatigue related features in high accuracy in that the eye sizes, the position of the driver to camera basically effects the measurement. The frequency of eye blinks may change on drivers even when they are not fatigue. Therefore, the camera-based solutions should calculate the adaptive features for decision making. In that study, we propose a driver adaptive eye blink-based fatigue detection system.

## 2. MATERIALS & METHODS

This study aims to develop a real-time driver fatigue detection system using eye blink features real-time.

## 2.1 Dataset

This study utilizes a proprietary dataset shared in collaboration with The University of Texas at Arlington (UTA or UT Arlington) to detect and analyze driver fatigue. The dataset, named UTA-RLLD, consists of videos from 60 different individuals, categorized into three different labels (0, 5, 10) designed to examine various aspects of driver fatigue [21]. These labels represent the levels of driver fatigue: "0" indicates non-drowsy, "5" indicates mildly drowsy, and "10" indicates drowsy drivers (Figure 4).

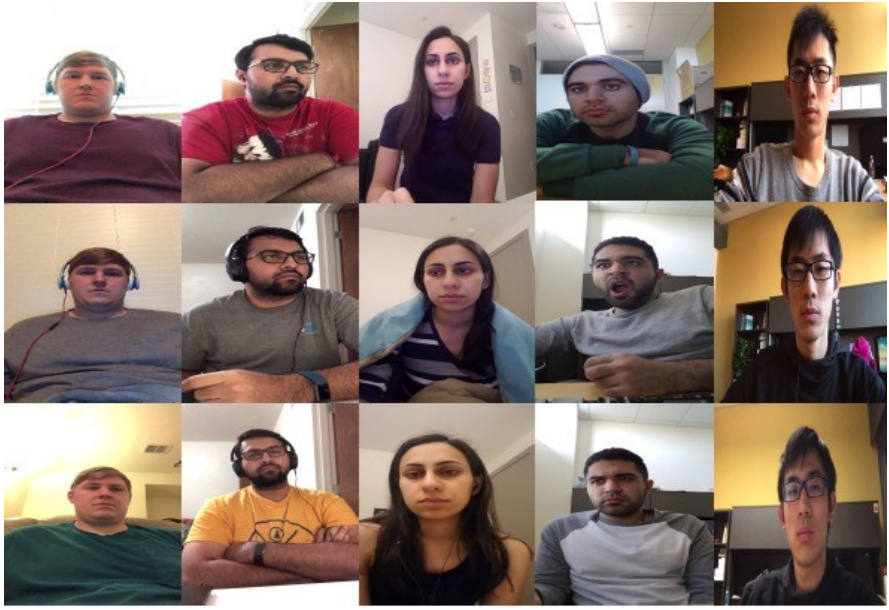

Figure 4. Example frames from the UTA-RLLD dataset showing not fatigued (top row), mildly fatigued (middle row), and highly fatigue (bottom row) states [21] of five different persons.

Each video was recorded for an average duration of 10 to 15 minutes, capturing various signs of fatigue such as facial expressions, head movements, and eye activities. These videos were recorded in a controlled environment simulating real-world conditions that drivers may encounter during long-term driving.

## 2.2 Detection of Eye Blink based Features

On average, a person blinks 10-15 times per minute, with each blink lasting between 100 and 400 milliseconds [14]. The average eye closure duration per minute is 1-6 seconds. The eye aspect ratio (EAR) basically changes during eye blinks. The landmark points (37-41) and (43-48) on Figure 3 give the perimeter of the right and left eyes. Using these points, the $EAR_{right}$, $EAR_{left}$ were calculated as

$$EAR_{right} = \frac{\|p_{38}-p_{42}\|+\|p_{39}-p_{41}\|}{2\|p_{37}-p_{40}\|}, EAR_{left} = \frac{\|p_{44}-p_{48}\|+\|p_{45}-p_{47}\|}{2\|p_{43}-p_{40}\|} \quad (1)$$

which changes when the eyes start closing and average of $EAR_{right}$ and $EAR_{left}$ are evaluated to represent EAR of driver. However, the EAR value may change when the driver looks at different direction during driving and any change on EAR should not be confused with eye blink. Therefore, a driver specific EAR threshold(s) must be defined to identify the time the blink starts and ends. Unlike other studies in literature, we assumed that drivers are not fatigue at the start of driving and the measurement taken at this period defines their regular blink and head movement behaviors. Therefore, we have evaluated real time features using the reference EAR and head position features. The reference EAR value ($EAR_{ref}$) was evaluated by getting the average of EAR during the first two minutes and experimentally 20% more than $EAR_{ref}$ was chosen as upper EAR threshold (thH) and 20% less than it was chosen as lower EAR threshold (thL).

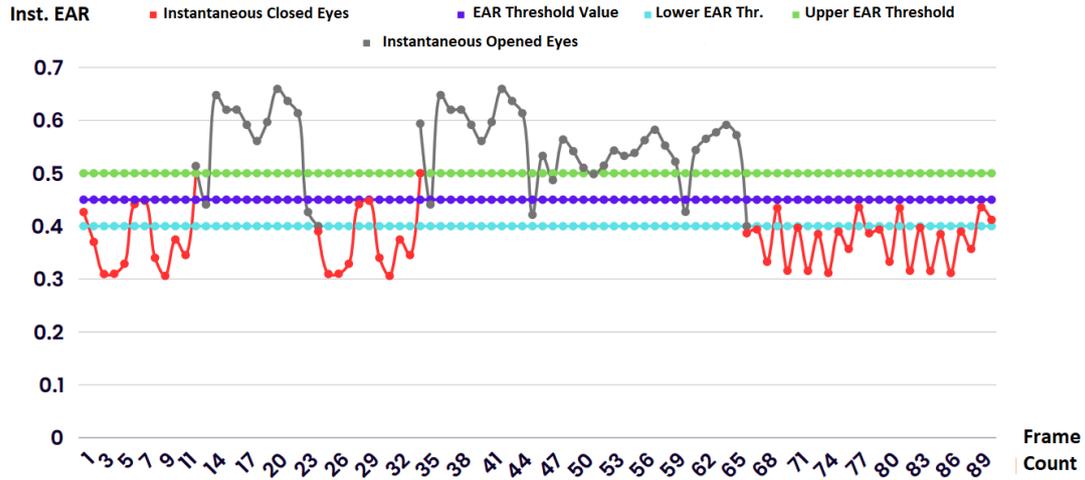

Figure 5. EAR at image frames and blinks.

Based on the determined thresholds, the transition from open to closed state begins only after crossing the lower threshold. The closed state continues until it reaches the upper threshold, and all the time in between is recorded as the eye being closed (red line on Figure 5). The minimum value during this process is recorded as the moment when the eye is fully closed. After reaching the upper threshold, the eye-open state is updated, and this process repeats until the next lower threshold is reached. The time between the blink starts and the next blink start forms a blink cycle, and a cycle includes regions where the eye is open and closed. We have evaluated the 13 different features for each blink cycle by considering the number of images frames within each region (open, closed) of blink cycle. As an example, the set of features for two different blink cycles are given at Table 1.

Table 1. Features of two different blinks.

| id | Features | Blink #1 | Blink# 2 |
|---|---|---|---|
| 1 | #of frames in open eye region | 3 | 105 |
| 2 | Minimum EAR in open eye region | 0.407 | 0.391 |
| 3 | Maximum EAR in open eye region | 0.53 | 0.5368 |
| 4 | Average EAR in open eye region | 0.473 | 0.5017 |
| 5 | Difference from reference open eye average (first 2 min) | -0.026 | -0.053 |
| 6 | #of frames in closed eye region | 158 | 1432 |
| 7 | Minimum EAR in closed eye region | 0.2469 | 0.1965 |
| 8 | Maximum EAR in closed eye region | 0.5289 | 0.5253 |

| 9 | Average EAR in closed eye region | 0.4798 | 0.4891 |
|---|---|---|---|
| 10 | Difference from reference close eye average (first 2 min) | -0.0312 | -0.0404 |
| 11 | #frames from open state to fully closed state | 2 | 440 |
| 12 | #frames from fully closed state to open state | 74 | 990 |
| 13 | Difference between frames from fully closed to open state | 137.8 | -778.2 |

The first 10 features (blink_set1) on Table 1 indicates EAR related features focused on closed and open eye region within a click cycle. However, the last three features (blink_set2) specifies the behavior of each blink which are expected to change if the drivers are fatigue.

## 2.3 Classification

Although the videos on dataset were labeled with three different level (non-drowsy, mid-drowsy and drowsy), a person may in one video may show other class's behavior at some instances. For example, at mid-drowsy videos, a person may show drowsy or non-drowsy behaviors at different instances. Even, in drowsy videos, there can be non-drowsy instances. To have a relatively consistent classification, we have performed binary classification using non-drowsy and drowsy videos. Using the generated feature sets (blink_set1 and blink_set2), we have used five different classifiers (decision tree, random forest, svm, k-nnn and logistic regression) and get the results in term of classification accuracy and F1-Score. The classifications were performed by using five-fold cross validation and the averaged results are presented.

## 3. EXPERIMENTAL RESULTS

The classifications were performed by using five-fold cross validation and the averaged results are presented for the blink_set1 and whole dataset (blink_set1 and blink_set2) to see the contribution of blinks inner characteristics to the blink cycle features (Table 2).

Table 2. Classification results

|  | blink_set1 | | blink_set1+b blink_set2 | |
|---|---|---|---|---|
|  | Accuracy (%) | F1-Score (%) | Accuracy (%) | F1-Score (%) |
| SVM | 77.89 | 71.85 | 87.53 | 82.74 |
| k-NN | 86.27 | 85.39 | 93.68 | 93.56 |
| Logistic Regression | 83.64 | 83.59 | 87.56 | 87.84 |
| Decision-Tree | 88.85 | 89.03 | 96.75 | 96.81 |
| Random Forest | 93.22 | 92.88 | 98.37 | 98.41 |

It is observed from Table 2 that for both datasets, the highest and the lowest accuracy and F1-Score were obtained by Random Forest and SVM classifier, respectively. The results also indicate the contribution of blink_set2 for drowsiness detection in that blink_set2 increased the accuracy and F1-Score in all the classifiers.

Although a video is labeled as a whole in dataset, we have tested a randomly selected video from non-drowsy and drowsy videos and showed the estimated classes for each instances (blinks) of the videos (Figure 6)

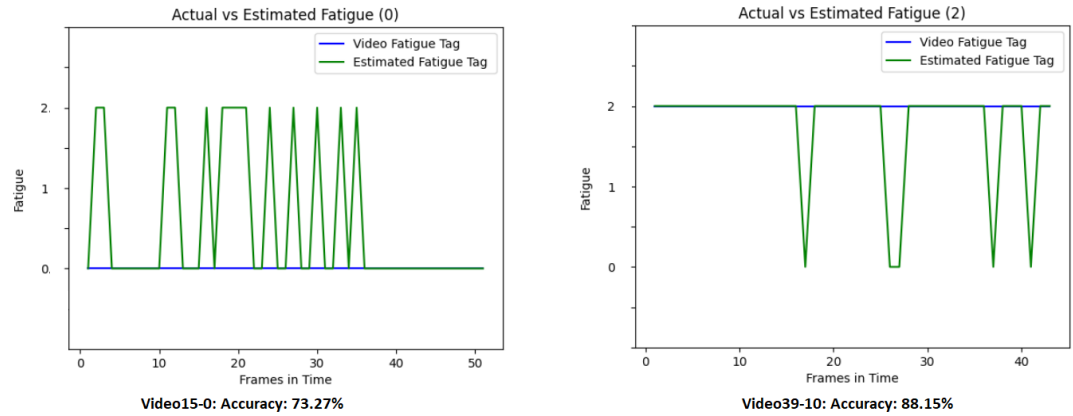

Figure 6: Prediction of a random selected non-fatigue (on left) and fatigue (on right) video. Label 0 (vertical axis) is non-fatigue, label 1 is fatigue estimation.

In is observed on Figure 6 that for the analysis of 50 frames, 73.23% of the time, the driver was detected as non-fatigue. Similarly, 88.15% of the time, a fatigue driver is detected as fatigue.

## 4. CONCLUSIONS

The proposed study focused on eye blink and generated eye aspect ratio (EAR) based features evaluated for each blink cycle which includes "eye open" and "eye closed/blink" region. To get representative features robust to drivers' physical eye characteristics and their position to camera, the first two minutes on the videos were regarded as non-drowsy period and adaptive thresholds were evaluated to identify the starting and ending points of the blinks. In addition to statistical features evaluated for "eye closed" and "eye open" region, three features were evaluated for blink characteristics such us the time of closing the eyes when the blink starts, the time of opening eyes after eyes are completely closed and their differences hypothesizing that the closing and opening of eyes in within blinks dramatically changes when the drivers are fatigue. Among the five different classifiers, the highest accuracy (93.22%) was obtained by random forest classifier when the blink_set 1 is user. However, the addition of in-blink features (blink_set2) the results were improved up to 10% percent. The result indicated the effect of blink characteristics for drowsiness detection. In addition to eye related features, the movement of head may also have rich information for drowsiness detection. Therefore, in future studies, it is planned to track head movement-based features together with blink-based for drowsiness detection.

## Authors

Ali Akin commenced his undergraduate studies in 2011 at the Izmir Institute of Technology in the Department of Computer Engineering and graduated in 2016. Following his graduation, Ali began his career at Accenture as a Java Developer, where he worked for three years. In 2019, he joined TÜBİTAK BİLGEM, serving as a software engineer for four years. In 2023, Ali transitioned to Definex, assuming the role of Lead Developer. Concurrent with his professional advancements, in 2021, he was admitted to the master's program in Computer Engineering at the Institute of Science at Gebze Technical University.

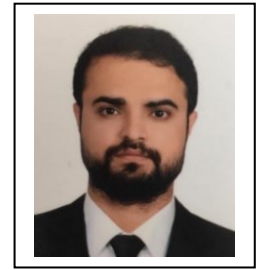

Habil Kalkan received the B.S. degree in Electrical and Electronic Engineering from Ege University,Turkey in 2002 and PhD Degree from Informatics Institute of Middle East Technical University in 2009. He has been a visiting researcher at the University of Minnesota and Kansas State University, USA in 2006. He has conducted post-doctoral research on medical image analysis at the Delft University of Technology, Netherland in 2011. Currently, He is a faculty member of Department of Computer Engineering of Gebze Technical University, TURKEY.  Dr. Kalkan's research fields are image processing, hyperspectral image processing for faulty detection, non invasive food safety application,  statistical data analysis and artificial intelligence. He coordinated several projects which were founded by national and international agencies including. Dr. Kalkan and his research group is currently working on usage of image processing and AI on real life applications.

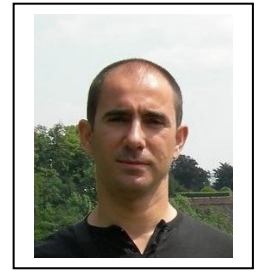